# ANNOTATED GUIDELINES AND BUILDING REFERENCE CORPUS FOR MYANMAR-ENGLISH WORD ALIGNMENT

Nway Nway Han and Aye Thida

AI Research Lab, University of Computer Studies, Mandalay, Myanmar.

*ABSTRACT*

*Reference corpus for word alignment is an important resource for developing and evaluating word alignment methods. For Myanmar-English language pairs, there is no reference corpus to evaluate the word alignment tasks. Therefore, we created the guidelines for Myanmar-English word alignment annotation between two languages over contrastive learning and built the Myanmar-English reference corpus consisting of verified alignments from Myanmar ALT of the Asian Language Treebank (ALT). This reference corpus contains confident labels sure (S) and possible (P) for word alignments which are used to test for the purpose of evaluation of the word alignments tasks. We discuss the most linking ambiguities to define consistent and systematic instructions to align manual words. We evaluated the results of annotators agreement using our reference corpus in terms of alignment error rate (AER) in word alignment tasks and discuss the words relationships in terms of BLEU scores.*

## 1. INTRODUCTION

A bilingual corpus aligned at the level of sentences or words is a precious resource for developing machine translation systems. Word alignment is a fundamental step in extracting translation information from bilingual corpus and determines which words and phrases are translations of each other in the original and translated sentence.

In most translation systems, translational correspondences are rather complex; for a language pair such as Myanmar and English that belong to the different word order languages. Myanmar is the Subject (S) Object (O) Verb (V) language and English the Subject (S) Verb (V) Object (O) language. Especially, Myanmar is the rich morphological language and also the free word-order language. Therefore, finding the word correspondences between Myanmar and English is quite unclear.

To evaluate the performance of the word alignment tasks, manually annotated reference corpora are needed. Moreover, to build the consistent reference corpora, annotated guidelines are also required for every annotator to resolve uncertain cases where correct corresponding items of the languages are difficult to find. Reference corpora in which sub-sentential translational correspondences are indicated manually also called Gold Standards that have been used as an objective means for testing word alignment systems.

However, there is no reference corpus and annotated guidelines for aligning between Myanmar and English languages. Therefore, to cover this problem, we proposed to build reference corpus and defined detailed annotation guidelines for the Myanmar-English by contrastive learning between two languages. As we mention above, Myanmar is the free word-order language and has a much more complex morphology than English. We will investigate the complex writing system of Myanmar in grammatical categories to improve the linking consistency between Myanmar-English language pairs.

## 2. RELATED WORK

There are several gold standards with manually annotated word alignments for other languages such as English- French [9], Dutch-English [1], English-Spanish [3], Chinese-Korean [2], Hindi-English [12] and Czech-English [4]. These are mainly to provide the performance of the word alignment systems.

In the annotation scheme of Arcade project [22] and the PLUG project [23], the alignments are explicitly allowed and translational correspondences are manually provided. For English-French word alignment system, [9] introduced sure and possible links to create a gold standard corpus for word alignment. Sure links were used for un-ambiguous alignments and possible links were used for ambiguous alignments (i.e. idiomatic expressions, free translations and missing function words). The approach was adopted for the English-Spanish language pair.

Blinker project, [19] created a detailed annotation style guide for the French-English language pairs which is provided to align translational divergences. The high degree of agreement between annotators indicates that the alignment task is feasible. In some projects [19, 21], the annotators were asked to explicitly mark those null-alignments, while in other projects[9] all unlinked words were considered to be null-alignments. In this paper, we create a reference corpus for Myanmar-English by using the full text word alignment and defined detailed annotation guidelines in the Section 5.

This paper is organized as follows. In Section 3 we select Text Selection for Annotated Corpus. Annotation scheme for confident labels are described in Section 4 and the guidelines for manual word alignment are motivated and demonstrated with contrastive analyses in Section 5. Then we show that how we build reference corpus and evaluate the reference corpus consistent with AGR statistics and AER scores of word alignment tasks in Section 6. Section 7 shows the evaluation results on Myanmar ALT corpus by using our reference corpus to discuss relationship between the word alignment and translation tasks. Finally, Section 8 contains conclusions and research plans for future work.

## 3. TEXT SELECTION

In some language pairs, parallel resources are developed in the form of parallel tree banks. Currently, free Myanmar- English parallel corpus is available at the Asian Language Treebank (ALT) [16]. This is a multilingual treebank consisting of 20,000 sentences from English Wikinews, and translating these sentences into the other six languages [14] which are closed to the English original as possible.

To create the consistent annotated guidelines for Myanmar Language with English, this guideline uses the verified word aligned Myanmar ALT data of the Asian Language Treebank. This corpus has word segmentation, part-of-speech tags, and syntactic analysis annotations, together with word alignment links among these languages.

However, there is no annotated with labels to evaluate the performance of the word alignment tasks. To address this problem, we are randomly collected 500 sentences pairs from 20,106 sentences are collected as the reference data and the maximum sentence length is 15(words). And then we annotated this information structure with confident labels by manually using annotated guideline which is described in section 5 to calculate more fine-grained evaluation measures.

## 4. ANNOTATION SCHEME FOR CONFIDENT LABELS

Word alignment with confident labels can be confusing or possibly with multiple possible choices for linking words with sure or possible link, so there exist ambiguities for human annotators. We use the link information defined in Myanmar ALT corpus to create a reference corpus that is consistent for the alignment of English-Myanmar bilingual texts.

In order to make it more general for many annotators to perform the alignment task with confident label, we proposed an annotation guidelines based on the EPPS guidelines [6]. The correspondence between two lexical units should involve on both sides as few words as possible but as many words as necessary, with the requirement that the linked words or groups connected have the same meaning. It is faired to systems which are based on either Blinker [2] or LinES [9] alignment guidelines. Sure (S) links were used for unambiguous alignments and possible (P) links were used for ambiguous alignments (i.e. idiomatic expressions, free translations and missing function words). All alignments are saved in NAACL format [4] and our test data guideline contains two types of links, sure(S) and possible (P) links. The use of sure and possible links in our guidelines is illustrated details in section 4.

In near future, our reference corpus is designed to provide a freely available resource for improving the word alignment of statistical machine translation.

## 5. MANUAL WORD ALIGNMENT WITH CONFIDENT LABELS ON THE MYANMAR ALT CORPUS

This section describes how we manually annotated the confident labels on the word alignment of the Myanmar ALT data and presents guidelines for Myanmar-English word alignment. Before defining a set of guidelines, we present a contrastive analysis of Myanmar and English and describe annotation guideline of the reference data with specific examples.

### 5.1. Contrastive Analysis of Myanmar and English

In the case of Myanmar to English word alignment, it is not easy to align the structure of the Myanmar with English grammar classifications and terminologies. The main reason is that the two languages belong to different in the word order. The sentence structure of the English is the subject (S) verb (V) object (O), but in Myanmar, it is SOV. The difference in this structure needs to be understood by understanding the subsequent explanation. English is a fixed position language; there is (relatively) orderly order. Myanmar is relatively free. Moreover, Myanmar language including the abundant morphology forms and the complexity of that form is more than English. The final ending and auxiliaries will cause great ambiguity in the annotation process. Although English has articles and other generic determiner, Myanmar does not have. Myanmar language has extra particles and postpositional markers that do not exactly fit nicely into the pattern of English grammar rules and classifications. These problems increase the level of uncertainty in the annotation process.

In this annotation guideline, we have discussed about six categories for Myanmar such as Noun, Verb, Punctuation, Paraphrases, Reduplication and Number with specific 20 examples of alignment. In each example, solid line ( ——— ) is used for sure (S) link and dash line ( - - - - ) is used for possible (P) link.

## 5.2. Nouns

In Myanmar Grammar [5], nouns are classified by the four types of meaning or representation, or by the four types of constructions. In next sub sections, we discuss these nouns categories with specific examples.

### 5.2.1. Proper Nouns

The proper nouns of various words including country name, organization, person name, city name, etc. which can be considered as indivisible or linked word to word. In "Figure 1", we use sure link to connect the corresponding parts of proper names.

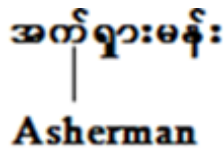

Figure 1.Example of Proper Nouns.

### 5.2.2. Compound Nouns

Some Myanmar compound nouns have no direct meaning in English but they correspond with noun phrase in English. We link the corresponding parts of Myanmar and English noun phrase by means of a sure link. Specific example of annotated word is illustrated in "Figure 2".

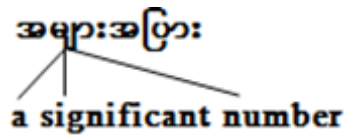

Figure 2.Example of Compound Nouns

### 5.2.3. Noun Phrase

Some Myanmar noun phrases correspond with direct meaning in English by one word. In "Figure 3", we link the corresponding subparts of the noun phrase in Myanmar with English word by means of a sure link and possible link for other words.

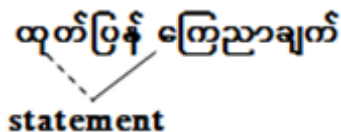

Figure 3. Example of Noun Phrase.

### 5.2.4. Common Nouns

Myanmar has no concept of definite English article marker (the). But most of common nouns in Myanmar Language correspond to the noun phrase together with the determiner (the) in English. We guided to use possible link to connect the determiner (the) to   the corresponding common noun in Myanmar word ("အစိုးရ"). We use sure link to connect the corresponding words ("အစိုးရ", Government) in both languages. Specific example of annotated word is illustrated in "Figure 4".

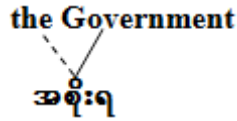

Figure 4. Example of Common Noun.

### 5.2.5. Singular Count Noun

For the cases of English article (a or an) which can be translated into a Myanmar singular count (တစ်ခု၊ တစ်ကောင်၊ တစ်ယောက်၊ တစ်စီး ... etc.). These words should be connected with an S link. Sometimes, there is no source word corresponds with English articles (a or an). In these cases, we use possible link to connect English articles (a or an) to the corresponding head noun in Myanmar. In "Figure 5", there is no source word corresponds with target word ("a"). Therefore, we use possible link between English article ("a") and the head noun of source word ("သင်္ဘော").

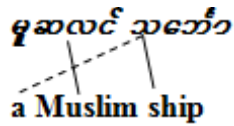

Figure 5. Example of Singular Count Noun.

### 5.2.6. Noun Related Particles

In Myanmar language, particles and post positional markers play critical roles for the nouns and pronouns in sentence structures. There are nine types of particles relating to the nouns in Myanmar. These noun-related particles include measure words or numerical classifiers, gender indicators, plural makers, words after numbers to show approximation, frequency or distribution, various support words to highlight the noun in a given situation, comparison words, exclamations words and words that describe all-inclusiveness. Those do not belong to a particular classification in English grammar. Therefore, we use a possible link to connect the Myanmar noun-related particles with the head noun in English.

In "Figure 6", the corresponding part of ("ရွာ" is "villages"), ("ငယ်" is "small"), ("၇" is "seven"). Therefore, we use sure link for that parts. However, the last word ("ရွာ" in "ရွာ ငယ် ၇ ရွာ") is noun related numerical particles and it has no corresponding words in English. We use possible link for that case

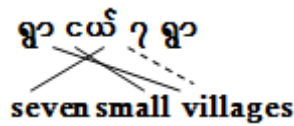

Figure 6. Example of Noun-related Particles.

### 5.2.7. Postpositional Markers

The postpositional markers (PPM) relating to the nouns and pronouns in Myanmar languages are often translated into a preposition or a possessive marker or no translation in English. We guided to annotate this usual practice is to connect the cases with the appropriate counterpart using an S

International Journal on Natural Language Computing (IJNLC) Vol.8, No.4, August 2019

link. If its counterpart is not explicitly present in target language then the case should be attached to its head noun of the source language and use possible link to connect the counterpart of head noun of the source language and phrasal mapping is to be done.

Specific example of annotated word is illustrated in "Figure 7". In this figure, "ပေါ်တူဂီ" means "Portugal", "မြို့" means "town", "သည်" and "ကို" in Myanmar words are postpositional markers. Since there is no relation to English, we use possible link for that relations

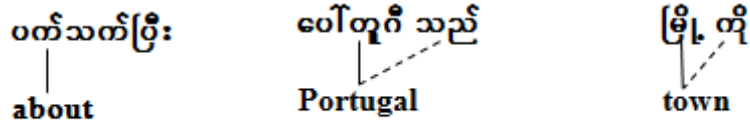

Figure 7. Examples of Postpositional Markers.

### 5.2.8. Genitive Case (Possessive case)

Myanmar genitive cases markers ("၏ / ရဲ့") are usually correspond to the possessives marker of English, typically as " *'s*" and " *of*". Sometimes these markers can be dealt within a manner similar to any other case ending( e.g. "သည်" ).In this case, the head noun of Myanmar should be mapped to the genitive of English (" *'s*" and " *of*") with possible link. If " *'s*" is glue with the head noun in target language then the case marker of Myanmar should be mapped with possible links to the head noun of the target language as shown in "Figure 8(a)". Otherwise, consider " *'s*" as separate word in target language and we use a sure link to connect the genitive case of Myanmar to its counterpart in "Figure 8(b)". Specific example of annotated word is illustrated in "Figure 8".

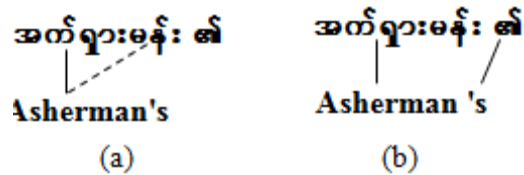

Figure 8. Example of Genitive Case in Noun.

### 5.3. Verbs

In the case of verb alignment for Myanmar-English sentences pairs, it is not easy to align the structure of the Myanmar language with English. The main reason is Myanmar language has extra particles and postpositional markers that do not exactly fit nicely into the pattern of English grammar rules and classifications. By the construction of the Myanmar sentence, verbs are also categorized into the three types: verbs that describe the act, verbs that show the quality, words that combine two actions [18].

### 5.3.1. Verbs that describe the act

For this category, if the source word has direct translation in the annotation process, the main verbs on both sides should be mapped with S links and the other auxiliary verbs or suffixes in English should be linked with Possible to the Myanmar main verb in the irrespective of their Mood, Aspect and Tense inflect forms. In "Figure 9", the main verbs of Myanmar-English are

International Journal on Natural Language Computing (IJNLC) Vol.8, No.4, August 2019

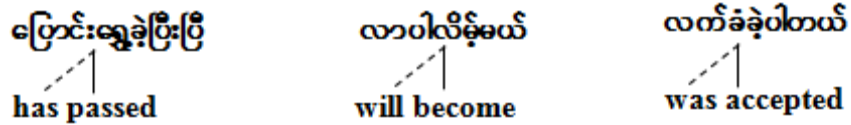

Figure 9. Example of Verbs that describe the act

"ပြောင်းရွှေ့ခဲ့ပြီးပြီ, passed ", "လာပါလိမ့်မယ်, become", "လက်ခံခဲ့ပါတယ်, accepted" and the auxiliary verbs or suffixes in English are "will", "has" and "was".

### 5.3.2. Verbs that show the quality

In this verb category in Myanmar which always corresponds to the combination of verb and adjective in English grammar as shown in "Figure 10". Therefore, we guided to annotate this word ("ထင်ရှားပြတ်သားသည်") in source language to the adjective ("clear") in the target language with sure link and the main verb ("is") in English with possible.

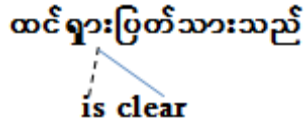

Figure 10. Example of Verbs that show the quality.

### 5.3.3. Verbs that combine two actions

For the case of the Myanmar verbs that can be followed by an another verb, it is possible to combine more than two verbs, and up to six verbs one after another is still a good writing style in the literary Myanmar sentences . In Figure 11, the two verbs of Myanmar are ("ဖြစ်" + "ပွား" = "ဖြစ်ပွား" and "ထုတ်" + "ဖော်" = "ထုတ်ဖော်", etc.). In the annotation process of this verb category in Myanmar, if its counterpart is not explicitly present in target language but has the same semantic meaning, groups of words should be linked together as shown in "Figure 11".

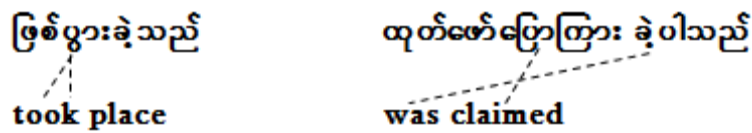

Figure. 11. Example of Verbs that combine two actions.

Among them, some usages are unheard of in the English language and these words are mostly used in Myanmar conversation sentences in which we can be embedded the subject noun or object noun or other words based on the situations. Therefore, we guided to annotate with the sure link between Myanmar words and its counterpart in English. We use possible link for other missing word. Specific example of annotated word is illustrated in "Figure 12".

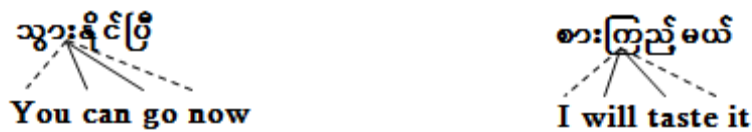

Fig. 12. Example of Verbs that combine two actions with embedded words.

### 5.3.4. Verbs that show the presence or existence of something

In Myanmar Language, verbs that show the presence or existence of something are always corresponding to the verb phrase (verb + preposition) in English. In this case, we use sure link to connect the verb phrase in English with the Myanmar verb that show the presence or existence. Specific example of annotated word is illustrated in "Figure 13".

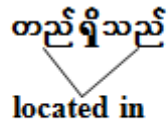

Figure. 13. Example of Verbs that show the presence or existence of something.

### 5.3.5. A support particle that immediately after the verb

In Myanmar, a support particle that immediately after the verb can be a particle, another verb, or an adverb. It is possible for a support particle to be followed by another support particle. Those words correspond to the following words in front of the verb "do" or "does" in English: can do, should do, seldom does, want to do, dare to do [18]. Therefore, we guided to annotate with the sure link between Myanmar words and its counterpart in English. We use possible link for other missing word. Specific example of annotated word is illustrated in "Figure 14".

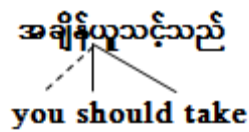

Figure. 14. Example of Verbs that show the support particle that immediately after the verb

### 5.3.6. Negative Verbs

In Myanmar, negative verbs (statements) are constructed into one word by combining the prefix particle 'မ' (argument), main verb and suffixes after verb to form negative imperatives and prohibitions. But in English grammar, the argument no or not can stand only in one word or appear together with other auxiliary words and main verb to form the negative sense. In these cases, we use sure link to connect the main verb and argument word in the English phrase with the negative verbs in Myanmar. For other auxiliary words in English, we use possible link to connect the negative verbs in Myanmar. Specific example of annotated word is illustrated in "Figure 15".

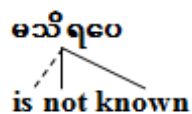

Figure. 15. Example of Negative Verbs

### 5.3.7. Interrogation Verbs

English has only a symbol of question mark ("?"), but in Myanmar language there are various types of question particles ("နည်း၊လော၊လား၊လဲ၊တုန်း") at the end of sentences. In Myanmar, these question particles are glued with the noun (ထမင်း), verb (ကျက်), postpositional marker (ပြီ, ကြောင့်), literary interrogative pronoun (အဘယ်, what, which, where), colloquial interrogative pronoun (ဘာ, what), colloquial adverb (ဘယ်လို, how) and literary pronoun (အသင်, you) base on

32

the relative information into one word. In English, the wh-clause can be used either as Nominal relative clause or question form. The difference between the two uses can only be made on context information. In some cases, non-interrogatory term of English (Do, Did, Does) can be used as auxiliary for interrogation [18]. Such terms are given sure links to their complement in English. We use possible link to connect the other auxiliary words in English interrogative to its counterpart in Myanmar. Specific example of annotated word is illustrated in "Figure 16".

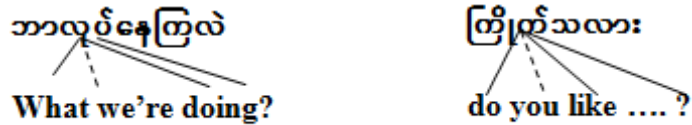

Figure. 16. Example of Interrogation verbs

### 5.4. Punctuation

In Myanmar Language, punctuation marks are ( ။ / ၊ ," ) whereas in English, punctuation are ( , . ," " : ; ! ? ). Moreover, Myanmar uses various types of question, emotions and moods particles at the end of sentences instead of question and exclamation marks (?,!) in English. If the marks of punctuation are correspondence on both sides, they should be connected to the S link. Otherwise, we use P link for the punctuation marks correspond to a word or another type of punctuation marks. Specific example of annotated word is illustrated in "Figure 17".

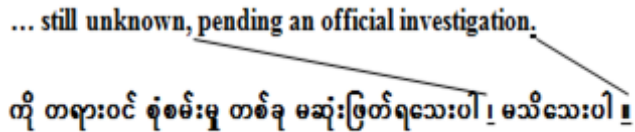

Figure. 17. Example of Punctuation

### 5.5. Paraphrases

If a meaning is paraphrased or expressed more explicitly in source or target sentence, use a possible link. If some words or word groups within the paraphrased section clearly correspond, mark these with a sure link. Specific example of annotated word is illustrated in "Figure 18".

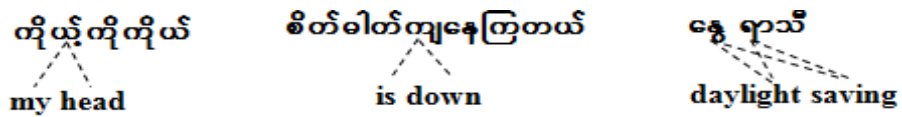

Figure. 18. Example of Paraphrases.

### 5.6. Reduplication

Reduplication adjectives in Myanmar either have an emphatic meaning or a distributive meaning. In these cases, we combined together those duplicated characters and mapped withsure link to its counterpart in English. Specific example of annotated word is illustrated in "Figure 19".

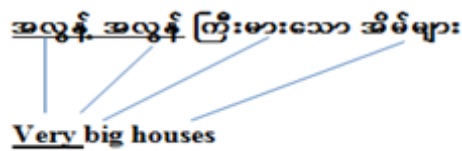

Figure. 19. Example of Reduplication

### 5.7. Number

In Myanmar language, number can be written in two forms: Myanmar Digit System and Myanmar Text System. In these cases, we guided to annotate these words to be corresponding words with sure(S) link and possible link (P) for other number particles. Specific example of annotated word is illustrated in "Figure 20".

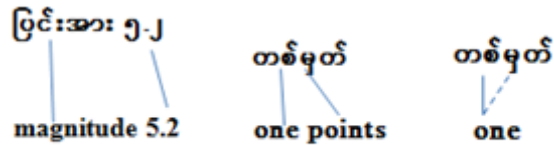

Figure. 20. Example of Number

## 6. ANNOTATOR AGREEMENT

In this section, we describe how we build the reference corpus using annotated guidelines mentioned above. Four annotators were given the task of drawing links between corresponding words and to give at least one link to all words in the sentence pair, using S and P according to the guidelines. All links from four annotators were added to the final reference alignment by using the following different options: Add all links to reference corpus if all annotators agreed Set possible (P) on the links if the count of possible (P) annotators is greater than the sure(S) annotators on that link. Set sure(S) on the links if the count of sure (S) annotators is greater than the possible (P) annotators on that link. Otherwise, set possible (P) on that link. The final reference corpus contains sure links (29%) and possible links (71%) over the 7486 links. Table 1 shows the ratio of each link type of each annotator and final reference corpus.

Table 1. Alignment Data

| Reference | %S | %P | Links |
|---|---|---|---|
| A1 | 29.37 | 70.63 | 7379 |
| A2 | 21.48 | 78.52 | 7422 |
| A3 | 23.06 | 76.94 | 7410 |
| A4 | 30.48 | 69.52 | 7416 |
| Final Reference | 29.16 | 70.83 | 7486 |

The alignment consistency between the four annotators is calculated as the AGR statistics AGR = $2*I/(A1+A2)$ where A1 and A2 is the sets of links created by the first and second annotator and I is the intersection of these sets. Table 2 shows the annotator agreements between four different annotators: between expert annotators E1 and E2, the expert's annotators and the novices after being exposed to the annotation guidelines.

International Journal on Natural Language Computing (IJNLC) Vol.8, No.4, August 2019Table 2. Agreement of each Annotator

| Reference | Alignment | AGR |
|---|---|---|
| E1 | E2 | **91.56** |
| E1 | N | 87.90 |
| E2 | N | 86.61 |
| E1 | N_E | 88.34 |
| E2 | N_E | 87.59 |

This system achieves 91.56 of AGR between two expert annotators and improvement ratio of E1 vs. N and E1 vs. NE is greater than E2 vs. N and E2 vs. NE. Based on this condition, the proposed guidelines create consistent and commendable annotation results. In addition, this guideline will help to develop consistent and systematic rules of word alignment for Myanmar-English language pair with different genres of corpus.

Table 3. Evaluation of each Expert Annotator with Final Reference Corpus

| Reference | Alignment | AER |
|---|---|---|
| R | E1 | 0.003197 |
| R | E2 | 0.003195 |

This system also evaluates each expert annotator with the final reference corpus (R) and describes the AER results in the following table. The result in Table 3 shows that each alignment error rate of 0.003 is nearly 0 occurred in these tests. This is due to the same perfect human alignments. The result in Table 3 also contains the very low error rates of test results in each annotation along with the other as reference.

## 7. EXPERIMENT ON MYANMAR ALT

In this section we use our reference corpus to make word alignment experiments on the Myanmar ALT parallel corpus. Our aim is to investigate the best word alignment tends to result in better translation quality.

To show the correlation between word alignment and translation matrix, we used four different methods (Intersection, Union, Grow-Diag (GD), and Grow-Diag-fial-and (GDFA)) of GIZA++ in both directions from Myanmar-English and English-Myanmar. Our analysis looked through a set of 500 bilingual sentence pairs which is a reference corpus.

For the experiments, we used standard phrase-based SMT system Moses [7] to conduct two different sizes of Myanmar-English ALT parallel corpora: small corpus contains 10K sentences pairs and large corpus contains 20K sentence pairs. With these observations, we were able to catch alignment errors and missed cases in Myanmar-English word alignment process. We report the performance of four different word alignments methods in terms of precision, recall and alignment error rate (AER) as defined by [9]. These three performance statistics are defined as follow,

35

Table 4. Evaluation for Small Corpus

| Align | Small Corpus | | | BLEU | |
|---|---|---|---|---|---|
| | P | R | AER | my-en | en-my |
| Intersect | 0.3145 | 0.1464 | 0.7971 | 6.57 | 3.74 |
| Union | 0.20466 | 0.27738 | 0.7673 | 5.25 | 3.46 |
| GD | 0.24382 | 0.22125 | 0.7672 | 7.14 | 4.24 |
| GDFA | 0.24249 | 0.23859 | **0.759** | **7.9** | **4.39** |

$$recall = \frac{|A \cap S|}{|S|} \qquad (1)$$

$$precision = \frac{|A \cap P|}{|A|} \qquad (2)$$

$$AER = 1 - \frac{|A \cap S| + |A \cap P|}{|S| + |A|} \qquad (3)$$

Where S indicates the set of annotated sure alignments, P indicates the set of annotated possible alignments, and A indicates the set of alignments produced by the model under the various sizes of training corpus. If sure alignment is not found, recall error will only occur and if not, precision error will only occur.

Alignments were evaluated against the 500 sentences in the reference corpus. Table. 4 and 5 show the quality of alignment and translation in precision, recall, AER and Bleu scores of each direction in Myanmar-English. Alignment with the highest accuracy takes the intersection (I) of the link from the two alignments, and the largest reconciliation alignment takes the union. Other heuristics methods such as grow-diag (GD) and grow-diag-final (GDF) are firstly created the intersect alignments and then add alignments from the union to increase alignment recall.

Table 5. Evaluation for Large Corpus

| Align | Small Corpus | | | BLEU | |
|---|---|---|---|---|---|
| | P | R | AER | my-en | en-my |
| Intersect | 0.315 | 0.156025 | 0.788707 | 8.25 | 4.8 |
| Union | 0.208113 | 0.279265 | 0.764107 | 7.24 | 4.63 |
| GD | 0.247602 | 0.227085 | 0.762466 | **9.1** | **5.2** |
| GDFA | 0.246712 | 0.244765 | **0.7542** | **9.51** | **5.57** |

Table 4 and Table 5 show that the lowest AER metric is likely to bring better translation quality when the system is trained with small and large corpora. In a system trained with small corpus in Table 4, GDFA alignment methods get the best (0.75 %) AER scores than the other different heuristics methods. On the other side in translation, this methods carried out the best the BLEU scores (7.9%) in Myanmar to English and (4.39 %) in English to Myanmar directions.

In Table 5, likewise in small corpus, this system gets the best AER (0.75%) in GDFA method and BLEU scores (9.51% and 5.57 %) in both directions which is trained on the large corpus.

However, there needs to make the optimized balance between precision and recall in two translational directions.

## 8. CONCLUSIONS

We presented a reference corpus for Myanmar-English word alignment which can be used to calculate the performance of the word alignment systems. We also described the alignment guidelines for manual annotation with trust tags (Sure and Possible) to build reference corpus for Myanmar-English word alignment systems. Moreover, we compared the results of different word alignment methods in a statistical machine translation system with our reference corpus to show the co-occurrence of the word alignment systems and the machine translation. We also intend to explore the relationship between alignment and translation by measuring other alignment characteristics that may affect the quality of translation, such as the types of words aligned and the number of discontinuities links.

### ACKNOWLEDGEMENTS

This work was annotated by Hnin Hnin, Win Thida Aung and Naw Hser Hser Gay who are members of AI Research Lab in University of Computer Studies, Mandalay, Myanmar.

### REFERENCES

[1] L. Macken, "An annotation scheme and Gold Standard for Dutch-English word alignment", In 7th conference on International Language Resources and Evaluation (LREC 2010) (pp. 3369-3374). European Language Resources Association (ELRA). J. Clerk Maxwell, A Treatise on Electricity and Magnetism, 3rd ed., vol. 2. Oxford: Clarendon, 1892, pp.68–73, 2010.

[2] J. Li, D.I. Kim and J.H. Lee, "Annotation Guidelines for Chinese-Korean Word Alignment", In LREC. May, 2008.

[3] P. Lambert, A. De Gispert, R. Banchs and J.B. Mariño, "Guidelines for word alignment evaluation and manual alignment", Language Resources and Evaluation, 39(4), pp.267-285, 2005.

[4] I. Kruijff-Korbayová, K., Chvátalová and O., Postolache , "Annotation Guidelines for Czech-English Word Alignment", In LREC , pp. 1256-1261, 2006.

[5] Y.K., Thu, W.P. Pa, M. Utiyama, A.M., Finch and E. Sumita, "Introducing the Asian Language Treebank (ALT)", In LREC, May, 2016.

[6] P. Koehn, "Statistical machine translation", Cambridge University Press, 2009.

[7] P. Koehn, H. Hoang, A. Birch, C. Callison-Burch, M. Federico, N. Bertoldi, B. Cowan, W. Shen, C. Moran, R. Zens and C. Dyer, 2007, June. Moses: Open source toolkit for statistical machine translation. In Proceedings of the 45th annual meeting of the association for computational linguistics companion volume proceedings of the demo and poster sessions (pp. 177-180).

[8] A., Fraser and D. Marcu , "Measuring word alignment quality for statistical machine translation", Computational Linguistics, 33(3), pp.293-303, 2007.

[9] F.J. Och and H. Ney, "A systematic comparison of various statistical alignment models", Computational linguistics, 29(1), pp.19-51. 2003.

[10] P. F. Brown, S. A. Della Pietra, V. J. Della Pietra, and R. L. Mercer, "The mathematics of statistical machine translation: Parameter estimation", Computational Linguistics, 19(2), pp.263–311. 1993.

[11] M. L. Commission. "Myanmar Thdda, Department of the Myanmar Language Commission", Ministry of Education, Union of Myanmar, 2005.

[12] R.K. Yadav and D. Gupta, "Annotation guidelines for Hindi-English word alignment", In 2010 International Conference on Asian Language Processing IEEE. pp. 293-296, December 2010.

**AUTHORS**

Nway Nway Han She received her B.C.Sc (Hons) and M.C.Sc degrees from University of Computer Studies, Mandalay, Myanmar. Currently, she is a candidate for the degree of Ph.D of Information Technology in University of Computer Studies, Mandalay in Myanmar. Her research interests include Natural Language Processing and machine translation.

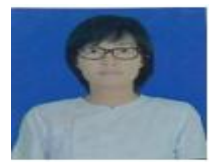

Aye Thida She is a Professor, at Faculty of Computer Science (Artificial Intelligence Lab), University of Computer Studies, Mandalay (UCSM), Myanmar. She is a leader of Natural Language Processing Project. Her research interests include High Performance Computing, Big Data Management and Artificial Intelligent. She is currently working NLP researches. She received B.Sc. (Hons) Maths degree from the Mandalay University, Myanmar and her M.I.Sc and Ph.D degrees in Information Technology from the University of Computer Studies, Yangon (UCSY), Myanmar.

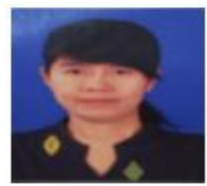